\documentclass[runningheads]{llncs}
\usepackage{tikz}
\usepackage{pgfplots}
\usepackage{amsmath}
\usepackage{tikz}
\usepackage{graphicx}
\usepackage{caption}
\usepackage{subcaption}
\usepackage{float}
\usepackage{indentfirst}
\begin{document}
\title{An Empirical Study of  Attention Networks for Semantic Segmentation}
\author{Hao Guo\inst{1}, Hongbiao Si\inst{2}, Guilin Jiang\inst{2}, Wei Zhang\inst{3}, Zhiyan Liu\inst{4},Xuanyi Zhu\inst{2},Xulong Zhang\inst{5},Yang Liu\inst{1}\thanks{Corresponding author:Yang Liu, liuyang87@hnchasing.com}}

\institute{Hunan Chasing Securities Co.,Ltd.
	\and Hunan Chasing Financial Holdings Co., Ltd. 
	\and Hunan Chasing Digital Technology Co., Ltd.
	\and Hunan Chasing Trust Co.,Ltd.
	\and Ping An Technology (Shenzhen) Co., Ltd.}
\authorrunning{Hao Guo and et al.}


	\maketitle              % typeset the header of the contribution

\begin{abstract}
 Semantic segmentation is a vital problem in computer vision. Recently, a common solution to semantic segmentation is the end-to-end convolution neural network, which is much more accurate than traditional methods.Recently, the decoders based on attention achieve state-of-the-art (SOTA) performance on various datasets. But these networks always are compared with the mIoU of previous SOTA networks to prove their superiority and ignore their characteristics without considering the computation complexity and precision in various categories, which is essential for engineering applications. Besides, the methods to analyze the FLOPs and memory are not consistent between different networks, which makes the comparison hard to be utilized. What's more,  various methods utilize attention in semantic segmentation, but the conclusion of these methods is lacking. This paper first conducts experiments to analyze their computation complexity and compare their performance. Then it summarizes suitable scenes for these networks and concludes key points that should be concerned when constructing an attention network. Last it points out some future directions of the attention network.\\
	
\end{abstract}
\section{Introduction}
Recently,  the human brain inspires the study of the attention mechanism. That is, the human brain can quickly select an area from vision signal to focus \cite{itti1998model}. Specifically, when observing an image, humans will learn the position where attention should be concentrated in the future by previous observation. Besides, humans always pay low attention to the surrounding area of an image, rather than reading all pixels of the whole image at one time and adjusting the focus over time \cite{hayhoe2005eye}. This mechanism includes hard attention and soft attention \cite{guo2022attention}. Hard attention uses a fixed matrix to focus on a certain part of the picture. However, it can not be updated and can not work for different samples \cite{xu2015show}. Therefore, almost all methods focus on soft attention, which utilizes a learnable weight matrix in the convolution layer.\par
Initially, attention is used for machine translation to capture long-range features. It helps the deep network to significantly improve its performance. The attention function projects  query,key and value to an output which is obtained by a weighted sum of the values, where the weight is computed by the corresponding query and key \cite{vaswani2017attention}. The input of an attention function includes the query, keys, and values, which are all vectors, and the output is also vectors. The attention function calculates the query's attention to the key, and self-attention is its attention to itself, meaning query and key are both obtained from the same input. And then it was utilized in encoder-decoder network in semantic segmentation.\par

However, these networks always are compared with the mIoU of previous SOTA methods to prove their superiority and ignore that different attention networks are suitable for different scenes, considering the computation complexity and precision in various categories, which is important for engineering applications. What's more, these methods do not follow a critical standard to compare their FLOPs. Therefore, the comparisons are hard to be utilized. Besides,  the conclusion of these attention networks is lacking. Therefore, this paper conducts an empirical study to analyze and summarize typical attention networks to guide future research.

\section{Related Work}
\subsubsection{Segmentation}
Semantic segmentation is a task which assigns a category to all pixels in a picture \cite{yuan2020multi}. Many methods have been developed to tackle  problems  ranging from automatic driving \cite{wang2018understanding}, virtual reality \cite{valenzuela2019efficient}, human-machine interaction \cite{sun2022Pre-Avatar}, ,scene understanding \cite{zhao2018psanet}, medical image
analysis \cite{ravanbakhsh2020human},etc. Recently, the common strategy to solve semantic segmentation problems is the deep structure network, which is much more accurate than traditional methods.\par
A common  architecture of a semantic segmentation network is widely regarded as an encoder network, followed by an end-to-end structure decoder network. The encoder is a pretrained backbone, such as Resnet \cite{he2016deep}. The function of the decoder is to map the feature semantic calculated by the encoder to the pixel space to make pixel-level classification. \par

\subsubsection{Attention networks}

\begin{figure}
	\centering
	\includegraphics[scale=0.4]{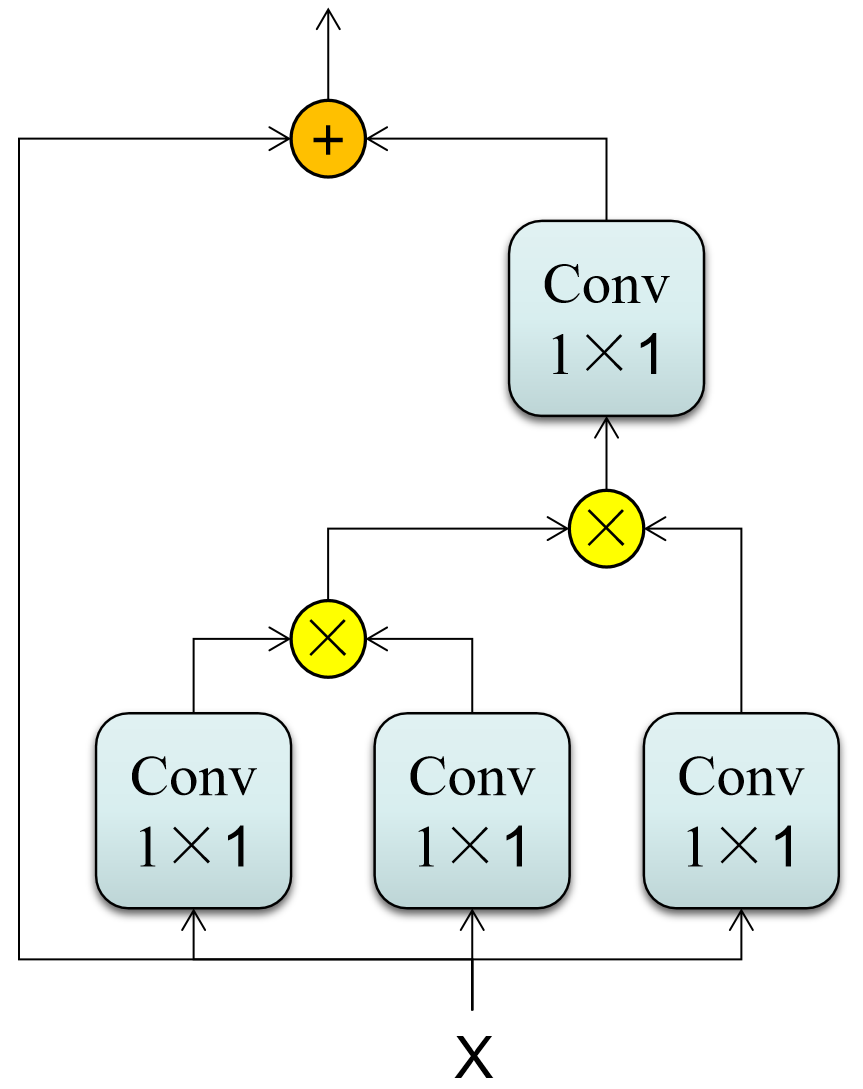}
	\captionsetup{font={large}}
	\caption{An overview of the Non-Local block.}
	\label{fig:nonlocal}
\end{figure}

 Non-Local Net bridges self-attention in machine translation and the Non-Local operations in computer vision. The self-attention networks introduced in the following part can be regarded as typical examples of this network.
The proposed Non-Local method calculates the interaction between every two points to obtain global dependencies without the limitation of convolution kernel size. Non-Local method is similar to utilize a convolution kernel with size of the feature map, which can maintain more contextual information. The structure of Non-Local block is shown in Fig.~\ref{fig:nonlocal}.\par

In the figure, $" \bigoplus "$ means element-wise sum, and $" \bigotimes "$ means matrix multiplication. Each operation includes softmax function. The blue boxes mean $1 \times1 $ convolutions.\par
\begin{figure}
	\centering
	\includegraphics[scale=0.4]{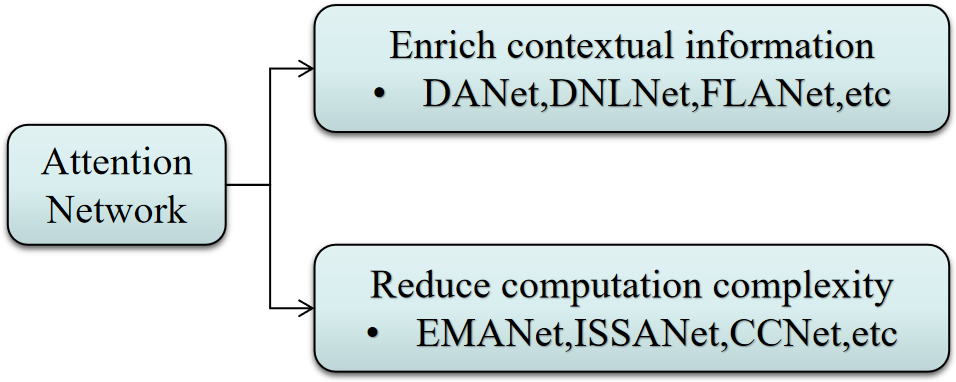}
	\captionsetup{font={large}}
	\caption{The classification of attention network.}
	\label{fig:type}
\end{figure}

Based on Non-Local Net, the following works can be divided into two types, as shown in Fig.~\ref{fig:type}. One is enriching the obtained contextual information, such as DNLNet, DANet, RecoNet, and FLANet, and the other is reducing the computation cost, such as EMANet, ISSANet, CCNet, and AttaNet. 

\subsection{Enrich  contextual information based methods}
	Non-Local Net captures global information in the spatial dimension. However,the channel dimension is overlooked. Compared with Non-Local Net, Dual Attention Network (DANet) \cite{fu2019dual} proposes two attention module to capture dependencies along the both dimension, respectively. Disentangled Non-Local Neural Networks (DNLNet) \cite{yin2020disentangled} decouples Non-Local operation to a whitened pairwise term to obtain the relationship between each pixel and a unary term to obtain the saliency of every pixel. However, these methods aim to construct a 2D similarity matrix to describe 3D context information. Therefore, a specific dimension is eliminated during the multiplication, which might damage the information representation. Tensor Low-Rank Reconstruction (RecoNet) \cite{chen2020tensor} splits feature map to tensors in height, width,and channel directions, and then reconstruct the context feature. Consequently, compared with the 2D similarity matrix, the reconstructed 3D context information can capture long-range features without eliminating any dimension. But the attention missing issue caused by matrix multiplication still exists. To solve this problem, Fully Attentional Network (FLANet) \cite{song2021fully} capture channel and spatial  attention in a single similarity map by proposed attention block.

\subsection{Reduce computation complexity based methods}
	Non-Local network enables each pixel to capture the global information. However, the self-attention mechanism produces a large attention map, which requires high spatial and temporal complexity. The bottleneck is that the attention map of each pixel needs to calculate the whole feature map. EMANet \cite{li2019expectation} utilizes the expectation maximization (EM) algorithm iteratively to obtain a set of compact ba ses and input them to the attention mechanism, which greatly reduces the complexity. Interlaced Sparse Self-Attention (ISSA) proposed an efficient scheme to factorize the computation of the dense matrix as the product of two sparse  matrices. CCNet \cite{huang2019ccnet}  introduces a recurrent criss-cross attention (RCCA) operation based on the criss-cross attention to solve this problem. The RCCA module can execute the criss-cross attention twice to capture the full-image dependencies.

\section{Experiment}

\subsection{Datasets}
	\begin{enumerate}
	\item[]\textbf{Cityscapes} \cite{cordts2016cityscapes} is a city scene dataset that can be used for semantic segmentation. It includes 20000 images with coarse annotations and 5000 images with fine pixel-level annotations and has 19 classes for semantic segmentation, and each image has a 1024 × 2048 resolution. The 5000 fine annotated images are divided into 2975, 500, and 1525 images for training, validation, and testing.
	\item[]\textbf{ADE20K} \cite{zhou2017scene} is a challenging scene parsing benchmark. This dataset contains 20K images for training and 2K images for validation. Images are densely labeled as 150 stuff/object categories and collected from different scenes with more scale variations.
	\end{enumerate}
\subsection{Implementation Details}
\subsubsection{Parameter Settings}
In the experiment, the learning rate is $1e^{-2}$, and models are trained for 80000 iterations  with batch size of 8. The encoder network is ResNet-101 which is a widely used backbone network.  The crop size is $769 \times 769$  for Cityscapes, and $520 \times 520$ for ADE20K. The poly learning rate policy is utilized which is $(1-iter/maxiter)^{0.9}$. And the weight decay coefficients and  momentum are  0.0005 and 0.9, which are common settings. And the synchronized batch normalization is utilized.  FLOPs, memory, and mIoU are used to evaluate runtime, memory, and precision, respectively.

\subsubsection{FLOPs Calculation}

However, there are some inconveniences by comparing FLOPs.
The first problem is that some papers do not provide the analysis of FLOPs, such as DANet and Non-Local Net. Besides, most papers do not provide the method for computing the FLOPs. Therefore their approaches to comparing the FLOPs might be different. Some papers use the open-source tools which follow the method in \cite{molchanov2016pruning}  to compute the  FLOPs in the whole network. However, this method is used to compute the FLOPs in the backbone, which does not involve matrix multiplication like $A\cdot B$. Therefore, the computation of matrix multiplication is overlooked, and the comparison is unfair.\par
 What's more, some papers analyze the whole network's FLOPs, and some compare the FLOPs in the attention module. And  Some networks, such as EMANet, calculate their FLOPs with input size of $513 \times 513 $, and some calculate their FLOPs with input size of $769 \times 769 $. These results are difficult to utilize for a fair comparison. 	\par
In this paper, following method \cite{molchanov2016pruning} which is also used by CCNet, we conduct experiments to calculate  FLOPs in the attention module with input size of $769 \times 769 $. The FLOPs of  convolutional kernels are defined as follows
\begin{equation}
	FLOPs=2HW(C_{in}K^{2}+1)C_{out}
\end{equation}
where $H$, $W$ and $C_{in} $are height, weight and input channel numbers, $K$ is the  size of kernel, and $C_{out}$ is the size of output channel.\par

\section{Analysis}
This section shows experiment results on attention networks and analyses  methods by which  they achieve their performance. Tab.\ref{flops} shows FLOPs of attention networks, and Tab.\ref{per-class} shows quantitative results on Cityscapes dataset.\\
\begin{table*}[h]

	\normalsize
		\centering
	\caption{Flops and Memory usage of typical attention  networks with input size $769 \times 769$.}
	\label{flops}
	{
		\begin{tabular}{l|cc}
			\hline
			Method &FLOPs(G) & Memory(M) \\
			\hline
			Denoised NL&8&428\\
			EMANet  &14 &100\\		
			CCNet&16&127\\
			FLANet & 19  &436\\
			ISANet &41 &141\\
			Non-Local Net &108&1411  \\
			DNLNet &108 &1450 \\
			DANet &113& 1462\\
			
			\hline
			
	\end{tabular}}
	\captionsetup{font={large}}

\end{table*}
\subsubsection{Non-Local Net}

Non-Local Net achieves scores of 78.5$\%$ mIoU on Cityscapes and 43.2$\%$ mIoU on ADE20K.
Compared with the baseline, which predicts results by the backbone, Non-Local Net gets 4$\%$ mIoU increment on Cityscapes validation set \cite{huang2019ccnet}. It demonstrates the proposed Non-Local operations directly capture global dependencies and significantly improve the segmentation performance. However, it also leads to high computation complexity. The FLOPs of Non-Local Net is much higher than its variations that are designed to reduce the computation cost.

\subsubsection{DANet}
DANet achieves scores of 80.6$\%$ mIoU on Cityscapes and 43.8$\%$ mIoU on ADE20K. Compared with Non-Local Net, the increment of computation complexity comes from the channel attention module. Compared with the attention map obtained by Non-Local Net, which size is $HW \times HW$,  the attention map generated by the channel attention module is $C \times C$, which is not a big number. Therefore, DANet does not increase high computation on Non-Local Net. Tab.\ref{flops} shows the FLOPs is slightly over Non-Local Net, but the accuracy is much higher. It demonstrates the proposed Dual Attention block successfully enriches the contextual information without cost too much computation.
\subsubsection{DNLNet}
DNLNet achieves scores of 79.4$\%$ mIoU on Cityscapes and 43.7$\%$ mIoU on ADE20K.	Compared with Non-Local Net, DNLNet splits the computation of Non-Local block into two terms, a whitened pairwise term and a unary term. The unary term tends to model salient boundaries and the whitened pairwise tends to obtains within-region relationships. Decoupling these two terms can enhance their learning. The the unary term is achieved by $1\times 1$ convolution, and whitened pairwise term is consist of the spatial attention module with the whitened operation. Therefore, the increment of computation complexity only comes from the unary term, which can be overlooked, and the computation complexity is similar to Non-Local Net.
\subsubsection{FLANet}

FLANet achieves scores of 82.1$\%$ mIoU on Cityscapes and 46.68$\%$ mIoU on ADE20K, which are the best results among these attention networks. Unlike DANet, which considers channel and spatial dimensions separately,  FLANet proposes Fully Attentional Block to encode channel and spatial attention in a single similarity map.
And the improved accuracy shows it is important to integrate both dimensions.	Besides, compared with DANet, FLANet utilizes global average pooling to obtain a global view in channel Non-Local mechanism. Therefore it takes less computation.

\subsubsection{Denoised Non-Local}

Denoised NL achieves scores of 81.1$\%$ mIoU on Cityscapes and 44.3$\%$ mIoU on ADE20K. Besides, it has the least computation cost among these attention networks. Denoted NL utilizes convolution layers to obtain a smaller input feature map, which dramatically reduces the computation cost. What's more,  different from DANet and FLANet, which enrich the contextual information by leading in the channel dimension,  Denoised NL enriches the neighbor information to reduce noises in the attention map. The experiment results demonstrate that the contextual information can be more precise without inter-class and intra-class noises.  

\subsubsection{EMANet}

EMANet achieves scores of 78.6$\%$ mIoU on Cityscapes and 43.3$\%$ mIoU on ADE20K. Compared with Non-Local Net, EMANet simplifies the input to a set of compact bases by EM algorithm and can converge in only three times iterations. Therefore, EMANet greatly reduces the complexity, the computation is much more efficient than Non-Local Net. What's more, the performance is slightly over  Non-Local Net on Cityscapes dataset. 

\subsubsection{ISANet}
ISANet achieves scores of 81.9$\%$ mIoU on Cityscapes and 43.5$\%$ mIoU on ADE20K.	Since ISANet factorizes the computation of the dense affinity matrix to the product of two sparse affinity matrices, the FLOPs of ISANet is much less than  Non-Local Net. Besides, the ISANet can adjust the group number of the sparse affinity matrices to achieve the best computation efficiency. In the experiment, we set the group number to 8 to minimize the computation cost.

\subsubsection{CCNet}
CCNet achieves scores of 78.8$\%$ mIoU on Cityscapes and 44.0$\%$ mIoU on ADE20K. Compared to Non-Local Net, which links each pixel with all pixels, CCNet links each element to the row and column where the element is located, which is called criss-cross attention. It can obtain the contextual information of all the pixels on its criss-cross path. CCNet can adjust the recurrent times of RCCA module. In the experiment, we set the recurrent number to two,  which achieves a good trade-off between speed and accuracy. Although indirectly obtained global dependencies cannot generate better results than direct calculation, CCNet utilizes the prior knowledge that the similarity in criss-cross path is more important. The RCCA module utilizes the similarity in criss-cross path twice to enhance the attention map. Therefore, the accuracy is better than Non-Local Net. \par

\begin{figure}
	\centering
	\includegraphics[scale=0.6]{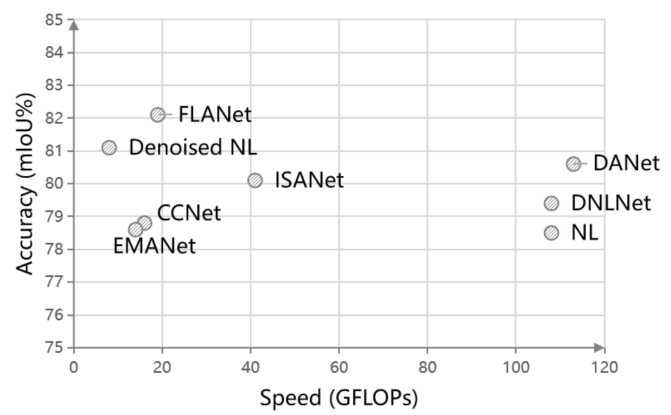}
	\captionsetup{font={large}}
	\caption{The comparison of accuracy and speed of attention networks. }
	\label{fig:compare}
\end{figure}

Tab.\ref{flops} shows the comparison of FLOPs and memory among different attention modules. The table shows that Denoised NL, FLANet, EMANet, and CCNet achieve the highest computation efficiency. They take much fewer GFLOPS than other networks. Besides, Fig.\ref{fig:compare} shows that FLANet achieves a good trade-off between accuracy and speed.\par

\begin{figure}
	\centering
	\includegraphics[scale=1.0]{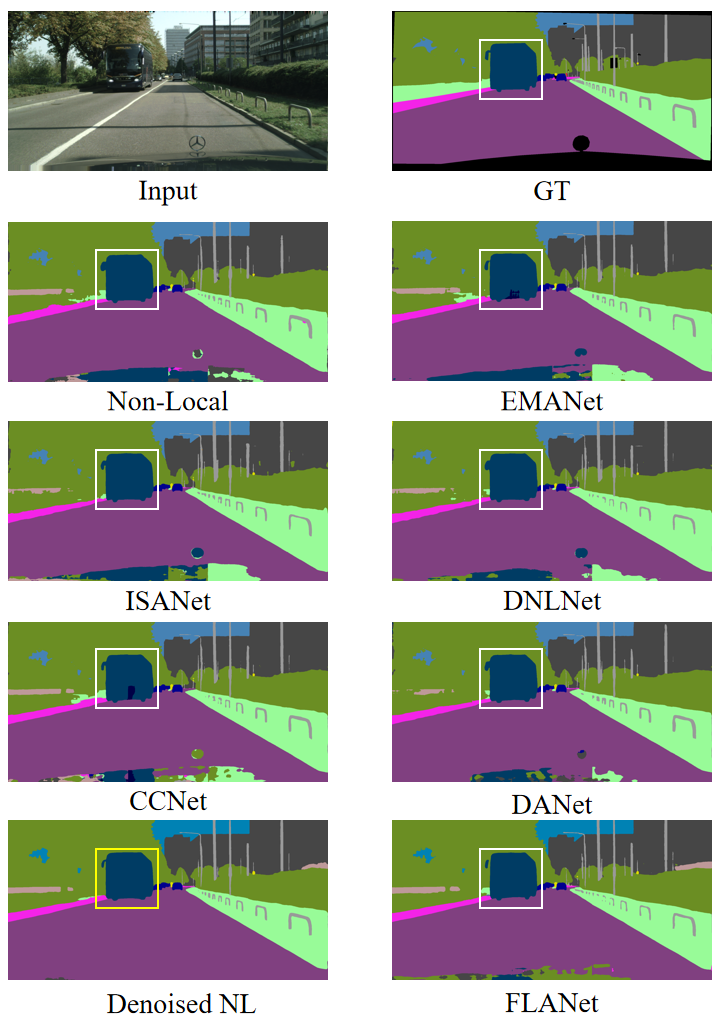}
	\caption{The qualitative results on Cityscapes. This figure shows the prediction of the class bus of attention networks. The white box is the class that should be focused on, and the yellow box is the prediction generated by the network which achieves the best accuracy in this class. }
	\label{fig:bus}
\end{figure}

\begin{figure}
	\centering
	\includegraphics[scale=1.0]{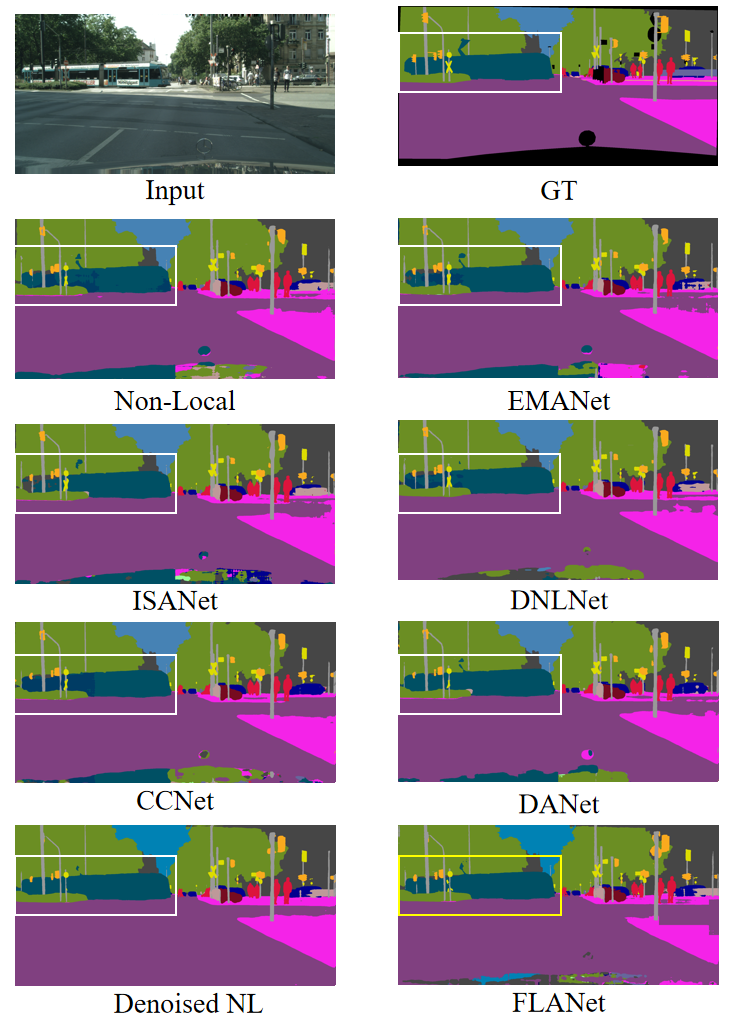}
	\caption{The qualitative results on Cityscapes. This figure shows the prediction of the class train of attention networks. The white box is the class that should be focused on, and the yellow box is the prediction generated by the network which achieves the best accuracy in this class. }
	\label{fig:train}
\end{figure}

This section also analyses the per-class mIoU of attention networks for the Cityscape dataset to find their characteristics. The quantitive results of experiments can be found in Tab.\ref{per-class}. From the table, all networks can get high scores for classes that are big objects and account for a large percentage of the dataset, such as roads and buildings. However, the results are not well for other classes that are hard to distinguish and recognize, such as walls, fences, and poles. Most classes such as Non-Local Net, EMANet, CCNet, and DNLNet cannot predict trains correctly. However, attention networks that enrich the contextual information, such as DANet and FLANet, can achieve high accuracy on trains with non-trivial increments. Besides, Denoised NL and FLANet achieve the best accuracy on per-class results, and the overall accuracy is much higher than other networks. For road,wall, build,wall,sidewalk,pole,sign,traffic light, terrain,person,sky,train,car,motorbike, and bike, FLANet achieves the highest accuracy. Besides, for vegetation, rider, truck, and bus, Denoised NL achieves the highest accuracy. This section conducts experiments and visualizes the prediction on the class bus and train, and the qualitative results can check  Fig.\ref{fig:bus} and Fig.\ref{fig:train}.\par 	

\begin{table*}
	\normalsize
	\centering
		\caption{Results on the Cityscapes val set.}
	\label{per-class}
	\resizebox{\textwidth}{!}{
		\begin{tabular}{l|ccccccccccccccccccc|c}
			\hline
			Method &  road &  swalk &  build  & wall  & fence &  pole  & tlight  & sign &  veg.  & terrain  & sky  & person  & rider  & car &  truck  & bus &  train &  mbike &  bike &  mIoU (\%)\\
			\hline
			Non-Local Net&98.1&84.8 & 93.2  &62.3 &62.2  &67.3& 74.1& 81.0& 92.5 &63.1& 95.1 &83.6&  66.6& 95.7& 82.9&  87.9& 73.9& 63.9 &79.1 &78.5\\
			EMANet &98.2 & 85.9& 93.3 &62.3 &63.0& 67.2 & 73.9 &81.2& 92.7& 64.1 &95.2&  84.1& 66.9& 95.8& 77.5& 84.0& 58.3& 65.2&  79.4&78.6\\
			CCNet&98.2& 85.4& 92.9&57.5&62.5& 67.0& 73.7& 81.1& 92.7& 63.0& 95.0& 84.0& 67.7& 95.7& 83.4& 86.7& 69.0& 62.2& 78.8&  78.8\\	
			DNLNet&98.1& 85.6& 93.1 &60.6& 64.0 &67.4 &73.8 &81.0& 92.6 &65.7 & 95.0 & 83.8& 66.5 & 95.0 &78.6 & 86.9 &72.1& 67.3  & 79.3&  79.4\\
			ISANet &98.2 &85.5 &93.0& 58.8 &63.8& 66.8 &74.4& 80.2  &92.6 &64.0 &95.2& 83.5& 66.0& 95.8 &84.5& 91.3  &79.0& 68.9& 79.3& 80.1\\
			DANet &98.2& 85.8& 93.4 &61.1& 65.9 &67.4 &74.0 &81.6& 92.8 &64.9 & 95.2 & 83.9& 67.1 & 95.7 &84.5 & 88.5 &84.0& 71.4  & 79.1&  80.6\\
			Denoised NL & 98.4&87.1 & 93.2 & 59.4 &64.5 &67.6 &73.0 &80.3 &93.1  &67.4 & 95.4 & 84.5 & 69.1 &95.6  & 84.6 &  \textbf{92.3}&  84.3 & 68.9  &79.3 &81.1 \\
			
			FLANet & 98.5&87.2 & 94.3 & 64.1 &67.5 &68.6 &78.9 &82.6 &93.0  &73.5 & 96.2 & 84.7 & 68.7 &96.6  & 83.0 &  91.7&  \textbf{86.2} & 74.3 & 80.0 & \textbf{82.1} \\
			\hline

	\end{tabular}}
	\captionsetup{font={large}}

\end{table*}

In Fig.\ref{fig:bus}, the yellow box indicates  Denoised NL achieves the highest performance on the class bus. Denoised NL captures clear edge information, and the prediction inside the bus is consistent. Compared with Denoised NL, EMANet and CCNet can capture clear edge information, but the prediction inside the bus shows an inconsistency. This observation shows why the mIoU is much lower than other networks.\par
In Fig.\ref{fig:train}, the results show  FLANet can make a much better prediction on the class train. Other networks cannot capture correct details in the train. For example, the inconsistency in the train shows the train was predicted as a wall or bus. And the tree on the left of the train is misclassified. Besides, the wire on the top cannot be predicted. These observations confirm why FLANet can achieve non-trivial increments on the class train.

\begin{table*}
	\normalsize
	\centering
		\caption{Results on the ADE20K val set}
	\label{ade}
	{
		\begin{tabular}{l|c}
			\hline
			Method & mIoU(\%)\\		
			\hline
			Non-Local Net &43.2\\
			EMANet  &43.3 \\
			ISANet&43.5 \\
			DNLNet&43.7\\
			DANet &43.8\\	
			CCNet&43.9\\
			Denoised NL& 44.1\\
			FLANet &  \textbf{44.2}\\
			\hline
			
	\end{tabular}}
	\captionsetup{font={large}}

\end{table*}
This section also conducts experiments on ADE20K datasets. Following previous works, it does not analyse the per-class mIoU. Since the ADE20K dataset contains 150 classes, it is pretty hard to analyse the characteristic of each class for all networks. Tab.\ref{ade} shows the overall performance by mIoU. The results show that FLANet and Denoised NL can also achieve the highest accuracy. 

Besides, Tab.\ref{test} shows the comparison of these attention networks with SOTA methods on Cityscapes test set. Denoised NL and FLANet also outperform most SOTA methods, except HSSA. HSSA is the latest work which get the best prediction by MaskFormer \cite{cheng2021per} which is transformer structure. Therefore, the computation cost of HSSA is much higher than FLANet. When HSSA utilizes ResNet-101 as backbone, the mIoU is 83.02\%, which is outperformed by FLANet.
In conclusion, for most segmentation tasks, FLANet is a good choice. It achieves the highest accuracy with high computation efficiency.\newpage

\begin{table*}
	\normalsize
	\centering
		\caption{The Comparison on  Cityscapes test set}
	\label{test}
	
	{
		\begin{tabular}{l|c}
			\hline
			Method & mIoU(\%)\\		
			\hline
			Non-Local Net&78.3\\
			EMANet &78.4 \\
			CCNet &78.6\\
			DNLNet&79.3\\			
			ISANet&79.9 \\
			DANet &80.4\\
			Segmenter \cite{strudel2021segmenter} &81.3 \\
			ACNet \cite{hu2019acnet} & 82.3\\
			GFF \cite{li2020gated} & 82.3\\
			SETR \cite{zheng2021rethinking} & 82.2\\
			OCR \cite{yuan2020object} &82.4\\
			DRANet \cite{hu2019acnet} & 82.9\\
			SegFormer \cite{xie2021segformer} &83.1\ \\
			Denoised NL& 83.5\\
			FLANet & 83.6\\
			HSSN \cite{li2022deep}  & \textbf{83.7}\\
			\hline
			
		\end{tabular}
	}
	\captionsetup{font={large}}

\end{table*}

\section{Conclusions and Future Works}
This paper first introduces the background and development of semantic segmentation and attention. Then it presents an overview of several attention networks. Besides, this paper conducts experiments to evaluate their performance by accuracy and speed. Last, it analyses these attention networks through qualitative results and quantitative results. \par
In conclusion, for most segmentation tasks, FLANet is a good choice. It achieves the highest accuracy with high computation efficiency. And Denoised NL, which has the fastest speed, can be utilized for the task that requires high computation efficiency.
Besides, to construct an attention network that can achieve high accuracy, we can enrich the contextual information in the attention mechanism from four aspects. First,  globally shared and unshared attention can be disentangled. Second, both channel-wise and spatial-wise attention are important. Third, when utilizing channel-wise and spatial-wise attention,  the attention missing issues should be avoided. Forth,  attention noises should be eliminated. To reduce the computational cost, we can consider two aspects. First, the input can be simplified. Besides, the process of obtaining the dense affinity matrix can be simplified.\par
This paper summarizes that FLANet and Denoised NL are proposed to enrich contextual information. However, FLANet proposes to encode channel attention and spatial attention in one attention map. Denoised NL proposes to encode neighbor information in the attention map. Both networks still lack contextual information from others. We may merge these two networks into one that can solve these two problems in future work.\par
Besides, the transformer-based methods achieve non-trivial increment on various tasks, including semantic segmentation. Since transformer structure is based on multi-head attention, similar to introduced attention networks, these networks also modify structure based on attention module. Therefore, 
if these attention networks can be transferred to a transformer structure, it is also valuable to be studied.

	%
	% the environments 'definition', 'lemma', 'proposition', 'corollary',
	% 'remark', and 'example' are defined in the LLNCS documentclass as well.
	%
	
	
	\bibliography{database}
	\bibliographystyle{splncs04}

\end{document}